\documentclass{article}
\PassOptionsToPackage{numbers,compress}{natbib}
\usepackage[preprint]{neurips_2026}

\usepackage[utf8]{inputenc}
\usepackage[T1]{fontenc}
\usepackage{hyperref}
\usepackage{url}
\usepackage{booktabs}
\usepackage{amsfonts}
\usepackage{amsmath}
\usepackage{amssymb}
\usepackage{nicefrac}
\usepackage{microtype}
\usepackage{xcolor}
\usepackage{multirow}
\usepackage{graphicx}
\usepackage{caption}
\usepackage{subcaption}
\usepackage{colortbl}
\usepackage{pifont}

\definecolor{best}{HTML}{FF4444}
\definecolor{second}{HTML}{4444FF}
\definecolor{lightgray}{HTML}{F0F0F0}

\title{PRISMR: Overcoming Parse Collapse in Multimodal Listwise Ranking via Parameterized Representation Internalization}

\author{%
  Hao Jiang\textsuperscript{1} \quad
  Xin Li\textsuperscript{1} \quad
  Annan Wang\textsuperscript{1} \quad
  Zhi Yang\textsuperscript{2} \\
  \textbf{Haoxiang Zhang\textsuperscript{2}} \quad
  \textbf{Yichi Zhang\textsuperscript{3}} \quad
  \textbf{Weisi Lin\textsuperscript{1}}\thanks{Corresponding author.} \\[6pt]
  \normalfont
  \textsuperscript{1}Nanyang Technological University \quad
  \textsuperscript{2}Peking University \quad
  \textsuperscript{3}Independent Researcher \\[4pt]
  {\small\texttt{\{jianghao907, lixin.1997.lixin, haoxiang.z.f, yichizhang0926\}@gmail.com}} \\
  {\small\texttt{annan001@e.ntu.edu.sg \quad zhiyang25@stu.pku.edu.cn \quad WSLin@ntu.edu.sg}}
}

\begin{document}

\maketitle

\begin{abstract}
Generative listwise ranking with Large Multimodal Models (LMMs) aims to capture global list context in a single forward pass, but its effectiveness degrades in long-context multimodal scenarios. We identify a recurring failure mode, \emph{parse collapse}, where the autoregressive decoder produces fluent yet incomplete rankings by silently omitting candidates and terminating early. This failure stems from limited context utilization rather than simple formatting mistakes, making prompt engineering and constrained decoding insufficient.
We propose \textbf{PRISMR} (\textbf{P}arameterized \textbf{R}epresentation \textbf{I}nternalization for \textbf{S}emantic \textbf{M}ultimodal \textbf{R}anking), a framework that replaces transient in-context list processing with parametric structural conditioning. PRISMR uses a lightweight hypernetwork to encode multimodal candidates in parallel and generate item-specific LoRA weights, which are synthesized into an instance-specific adapter for a LMM. This paradigm enables more robust internalization of list structure while preserving the base model. We further introduce a large-scale multimodal review-ranking benchmark for evaluation. Experiments demonstrate that PRISMR substantially reduces parse collapse, improves listwise ranking performance, and transfers effectively across domains and instruction-tuned backbones.
\end{abstract}

\section{Introduction}
\label{sec:intro}

Large Multimodal Models (LMMs)~\cite{achiam2023gpt,team2023gemini,yang2025qwen3} have demonstrated remarkable capabilities in reasoning across intertwined text and visual contexts. As these models are increasingly deployed in Learning-to-Rank (LTR) scenarios, existing ranking paradigms face a persistent trade-off in long-context settings ~\cite{jiang2026rlpo}. Traditionally, pointwise scoring is computationally efficient but fundamentally fails to account for list-level interactions, leading to miscalibrated top-$k$
rankings~\cite{liu2025llm4ranking,gera2025justrank}. Pairwise ranking captures relative preferences but entails an intractable $\mathcal{O}(N^2)$ computational complexity during inference. While generative listwise ranking~\cite{gupta2025scalable,liu2025lipo,cai2025k,wu2025context,reddy2024first,ren2025self,liu2025coranking}
 theoretically resolves these inefficiencies by evaluating the entire candidate list to leverage global context, it becomes highly unstable and computationally expensive as candidate lists grow. 

\begin{figure}[ht]
\centering
\includegraphics[width=\columnwidth]{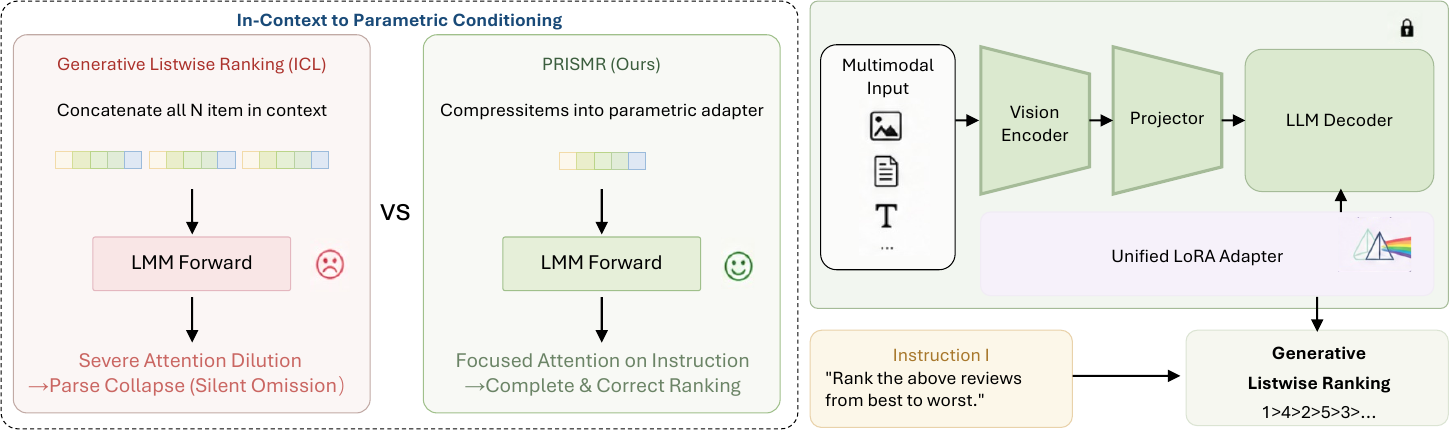}
\caption{
Conventional generative listwise ranking feeds all multimodal candidates into the decoder context, causing attention dilution and parse collapse as the list grows. 
}
\label{fig:teaser}
\end{figure}

This instability manifests as a severe generative fragility when deploying LMMs for multimodal listwise ranking. As seen in Fig.~\ref{fig:teaser}, When processing long, multimodal contexts---where extensive text interleaves with dense visual features---LMMs suffer from exacerbated attention dilution and the ``lost in the middle'' phenomenon~\cite{liu2024lost}. Consequently, LMMs frequently experience a catastrophic \textit{parse collapse}. The autoregressive generation process fails to adhere to strict structural constraints, resulting in severe hallucinations, missing candidates, and unparseable conversational outputs rather than the requested ranked lists. Existing adaptation approaches fall short of addressing this bottleneck. Standard Supervised Fine-Tuning (SFT) and Direct Preference Optimization (DPO) attempt to enforce output formats by permanently altering global weights, but they struggle to dynamically adapt to highly variable, instance-level multimodal contexts. Conversely, traditional Context Distillation (CD) methods require computationally prohibitive per-prompt gradient updates, rendering them unsuitable for real-time, large-scale ranking systems.

To overcome the inherent fragility of listwise generation and the inefficiency of current adaptation methods, we propose \textbf{PRISMR} (\textbf{P}arameterized \textbf{R}epresentation \textbf{I}nternalization for \textbf{S}emantic \textbf{M}ultimodal \textbf{R}anking). Shifting the paradigm from traditional in-context prompt processing to parametric structural conditioning, PRISMR leverages a lightweight hypernetwork to instantly encode both the complex system instructions and the rich multimodal candidate contexts (review content, item titles, and images) into a Low-Rank Adaptation (LoRA) module via a single, feed-forward pass. By projecting the cross-modal interactions and ranking-task constraints into a low-rank weight delta prior to decoding, PRISMR substantially reduces the in-context burden that drives parse collapse. The generated LoRA acts as a strong instance-specific structural prior during decoding, empirically pushing per-slot parse rate above $99.9\,\%$ in our experiments while retaining the base model's generalized world knowledge and zero-shot capabilities.

Extensive experiments demonstrate that PRISMR establishes a new state-of-the-art for multimodal listwise ranking. Our main contributions are summarized as follows:
\begin{itemize}

    \item \textbf{A length-adaptive PRISMR framework for multimodal listwise ranking.}
    We propose PRISMR, a parametric internalization framework that maps a long multimodal listwise instance into a feed-forward LoRA update, thereby moving candidate information from fragile prompt tokens into structured parameter space. A single trained hypernetwork supports two zero-cost test-time synthesis modes: $\alpha$-mode ($\sum_i B_iA_i$), which provides higher in-distribution capacity, and $\beta$-mode ($\tfrac{1}{N}\sum_i B_iA_i$), which improves robustness under length extrapolation. 

    \item \textbf{A multimodal listwise review-ranking benchmark.}
    We construct a domain-specific multimodal review-ranking benchmark from Amazon Reviews 2023~\citep{hou2024bridging}, where each example contains multiple reviews of the same product with titles, textual content, and user-uploaded images. The benchmark provides listwise supervision over review quality and will be publicly released to support research on long-context multimodal ranking.

    \item \textbf{A systematic analysis of parse collapse and ranking quality.}
    We identify parse collapse as a dominant failure mode of generative LMM listwise ranking, where models silently omit candidates or fail to produce valid rankings.  We characterize its dependence on list length and image density, and show that PRISMR improves format reliability.

\end{itemize}

\section{Related Work}
\label{sec:related_work}

\subsection{Multimodal Listwise Ranking and Generative Fragility}
Generative Large Language Models (LLMs) have reshaped Information Retrieval (IR) and Learning-to-Rank (LTR). While pointwise methods scale linearly but ignore list-level dynamics, generative listwise approaches like RankGPT~\citep{sun2023chatgpt} evaluate entire candidate lists to leverage global context. To directly optimize listwise metrics, recent works propose differentiable surrogate losses (e.g., diffNDCG~\citep{qiu2022large}) and Permutative Preference Alignment (PPA)~\citep{zhao2025permutative}. However, extending listwise paradigms to Large Multimodal Models (LMMs) introduces severe generative fragility. Processing long sequences of interleaved text and dense visual features exacerbates attention dilution and the ``lost in the middle'' phenomenon~\citep{liu2024lost}, triggering a ``parse collapse'' where models fail to respect output formats. While standard Supervised Fine-Tuning (SFT), Direct Preference Optimization~\citep{rafailov2023direct}, and reinforcement learning (e.g., GRPO~\citep{shao2024deepseekmath}) mitigate this on static distributions, they struggle to adapt to highly variable, instance-level multimodal contexts dynamically without catastrophic forgetting or reasoning overhead. Logits-only listwise rerankers like FIRST~\citep{reddy2024first} and RankZephyr~\citep{pradeep2023rankzephyr} sidestep generation failures but inherit the same long-context attention pressure as $N$ scales.

\subsection{Prompt Internalisation and Context Distillation}
To alleviate the computational bottleneck and attention dilution of long prompts, prior work explores Context Distillation (CD) and prompt compression. Compression techniques like LLMLingua~\citep{jiang2023llmlingua} and LLMLingua-2~\citep{pan2024llmlingua} drop tokens via information entropy, but inherently lose information and do not bypass context-window limits; we empirically confirm in Section~\ref{subsec:exp_listwise} that token-budget compression does \emph{not} alleviate parse collapse. Gisting~\citep{mu2023gisting} learns special ``gist'' tokens that compress an instruction prompt into a small constant number of soft tokens, while Snell-CD~\citep{snell2022contextdistillation} amortises a long context into model parameters by token-level distillation from a context-augmented teacher. Generative Prompt Internalisation (GenPI)~\citep{shin2025generative} jointly distils both the teacher's outputs and the prompt content. All three target a \emph{single} long context, whereas multimodal listwise ranking presents $N$ short multimodal items and requires the parametric encoding to be aware of relative ordering across the $N$ items.

\subsection{Hypernetwork-based PEFT}
Our work builds on hypernetwork-based PEFT~\citep{hu2022lora}: a small network produces task-specific weights conditioned on auxiliary information. HyperTuning~\citep{phang2023hypertuning} predicts soft prompts for a frozen language model conditioned on a few-shot description; HINT~\citep{ivison2023hint} predicts adapter parameters from natural-language instructions for zero/few-shot task generalisation; Text-to-LoRA~\citep{charakorn2025text} predicts LoRA weights from a task description for zero-shot adaptation; Doc-to-LoRA~\citep{charakorn2026doc} uses a Perceiver-based architecture to internalise a single long document into a LoRA adapter in one feed-forward pass. PRISMR is the listwise multimodal counterpart of this line of work: a single global hypernetwork encodes each of $N$ multimodal candidates into a per-item LoRA, and we explicitly study how to combine the $N$ adapters into one composite increment $\Delta W$. We derive closed-form rank-concatenation and mean-pooling operators, revealing a capacity and length-robustness trade-off unique to listwise hypernetwork PEFT.

\section{Method}
\label{sec:method}

\begin{figure}[ht]
\centering
\includegraphics[width=\columnwidth]{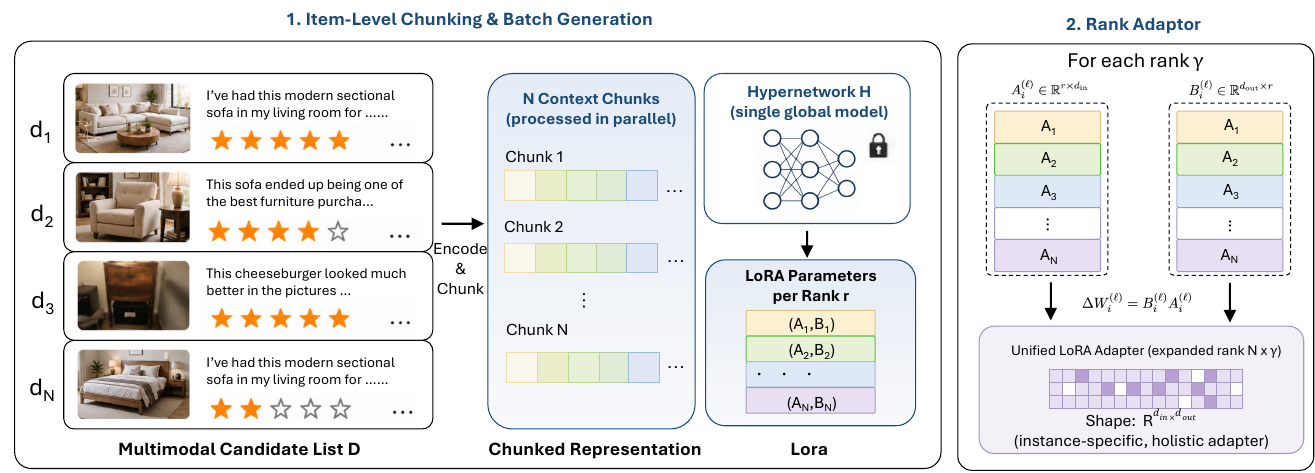}
\caption{Overview of PRISMR. A shared hypernetwork $\mathcal{H}_\phi$ maps each multimodal candidate $d_i$ to an item-specific LoRA adapter $(A_i,B_i)$. The $N$ adapters are combined by one of two zero-cost synthesis modes of the same checkpoint---rank-dimension concatenation ($\alpha$-mode) or mean pooling ($\beta$-mode)---yielding a composite increment $\Delta W$ mounted on the frozen LMM for listwise decoding. PRISMR uses $\alpha$-mode for $N\!\le\!50$ and $\beta$-mode for $N\!>\!50$ by default.}
\label{fig:overall}
\end{figure}

We formalize multimodal listwise ranking and present \textbf{PRISMR}, a framework that replaces long in-context list conditioning with instance-specific parametric conditioning. Instead of feeding all multimodal candidates into the decoder prompt, PRISMR uses a shared hypernetwork to encode each candidate into a low-rank adapter. The resulting adapters are then synthesized into a single weight increment mounted on a frozen LMM for one-shot listwise decoding. Figure~\ref{fig:overall} illustrates the overall architecture.

\subsection{Multimodal listwise ranking and parse collapse}
\label{subsec:problem_formulation}

Let $\mathcal{D}=\{d_1,\ldots,d_N\}$ denote a list of multimodal candidates, where each candidate $d_i=(t_i,v_i)$ consists of text $t_i$ and visual inputs $v_i$. Given an instruction $I$, a frozen LMM $f_\theta$ generates a target sequence $y=(y_1,\ldots,y_T)$ autoregressively:
\begin{equation}
P_\theta(y\mid c)=\prod_{t=1}^{T}P_\theta(y_t\mid y_{<t},c),
\label{eq:ar}
\end{equation}
where $c$ is the conditioning context.

In generative listwise scoring, the target output is a structured sequence
$y^\star=(\langle 1:s_1\rangle,\ldots,\langle N:s_N\rangle)$,
where $s_i\in[1,10]$ is the teacher score for candidate $i$. The generated scores induce a ranking $\pi$ via a deterministic monotone mapping, and evaluation is performed using NDCG@K against the teacher ordering $\pi^\star=\operatorname{argsort}_i(-s_i)$.

A standard in-context ranker conditions on the full multimodal list:
\begin{equation}
c_{\mathrm{ICL}}=[I;t_1,v_1;\ldots;t_N,v_N],
\qquad
|c_{\mathrm{ICL}}|=L_I+N L_d,
\label{eq:context_icl}
\end{equation}
where $L_d$ denotes the average candidate length. As $N$ grows, the decoder must attend over increasingly long multimodal contexts while also maintaining a strict output schema. Empirically, this leads to \emph{parse collapse}: the model produces fluent but incomplete outputs, silently omitting candidates or terminating early. 
PRISMR addresses this failure mode by moving candidate information from the decoder context into instance-specific parameters.

\subsection{Parametric internalization with a global hypernetwork}
\label{subsec:hypernet}

PRISMR keeps the base LMM frozen and introduces a shared hypernetwork $\mathcal{H}_\phi$. For each candidate $d_i$, the hypernetwork predicts low-rank LoRA factors for selected projections in the LMM:
\begin{equation}
\mathcal{H}_\phi(d_i)
=
\left\{(A_i^{(\ell)},B_i^{(\ell)})\right\}_{\ell=1}^{L},
\qquad
A_i^{(\ell)}\in\mathbb{R}^{r\times d_{\mathrm{in}}},
\quad
B_i^{(\ell)}\in\mathbb{R}^{d_{\mathrm{out}}\times r}.
\label{eq:hypernet}
\end{equation}
The induced item-specific increment at layer $\ell$ is
\begin{equation}
\Delta W_i^{(\ell)}=B_i^{(\ell)}A_i^{(\ell)}.
\end{equation}
The same hypernetwork is shared across all candidates and all examples. Since candidates are encoded independently, their adapters can be generated in parallel. 

\subsection{Adapter synthesis}
\label{subsec:synthesis}

The $N$ item-specific adapters are combined into a single increment $\Delta W$ before decoding. PRISMR exposes two zero-cost test-time synthesis modes of the \emph{same} trained hypernetwork. The first mode, denoted $\alpha$-mode, concatenates the LoRA factors along the rank dimension, with $A_\alpha=[A_1;\ldots;A_N]$ and $B_\alpha=[B_1,\ldots,B_N]$, yielding
\begin{equation}
\Delta W_\alpha
=
B_\alpha A_\alpha
=
\sum_{i=1}^{N} B_iA_i .
\label{eq:alpha_dw}
\end{equation}
This preserves item-specific subspaces and allows the effective rank to scale with the list size, i.e., $\mathrm{rank}(\Delta W_\alpha)\leq Nr$.

The second mode, denoted $\beta$-mode, averages item-level updates:
\begin{equation}
\Delta W_\beta
=
\frac{1}{N}\sum_{i=1}^{N} B_iA_i .
\label{eq:beta_dw}
\end{equation}
In contrast to $\alpha$-mode, mean pooling keeps the update scale and effective rank bounded as $N$ grows, with $\mathrm{rank}(\Delta W_\beta)\leq r$.

The two modes therefore induce a capacity--robustness trade-off within a single PRISMR checkpoint. $\alpha$-mode offers higher in-regime capacity because its effective rank grows with $N$, but for much longer lists its update norm can also grow with the number of candidates:
\begin{equation}
\|\Delta W_\alpha\|_2
\leq
\sum_{i=1}^{N}\|B_iA_i\|_2 .
\label{eq:alpha_norm}
\end{equation}
This may shift the adapted model outside the range observed during training. $\beta$-mode instead normalizes the aggregate update, making it more stable under length extrapolation. Crucially, both modes are obtained from a \emph{single} trained hypernetwork at zero additional training cost; PRISMR therefore adopts a simple length-adaptive rule by default: \textbf{use $\alpha$-mode for $N\!\le\!50$ (in-regime) and switch to $\beta$-mode for $N\!>\!50$ (extrapolation)}. Unless stated otherwise, ``PRISMR'' in the experiments below refers to this single checkpoint with the $N\!=\!50$ mode-switch threshold.

\subsection{PRISMR Loss}
\label{subsec:meta_training}

Training examples $(\mathcal{D},y^\star)$ are obtained from a frontier teacher model, which emits one structured score line per candidate. We discard trajectories with missing candidates, invalid indices, or malformed score lines. During training, the base LMM remains frozen, and only the hypernetwork parameters $\phi$ are updated.

We first optimize token-level distillation with the negative log-likelihood of the teacher sequence:
\begin{equation}
\mathcal{L}_{\mathrm{NLL}}(\phi)
=
-\sum_{t=1}^{T}
\log
P_{\theta+\Delta W(\phi)}
\left(
y_t^\star
\mid
y_{<t}^\star,c_{\mathrm{trig}}
\right).
\label{eq:l_nll}
\end{equation}
This objective encourages the adapted model to reproduce the teacher's structured scoring output from the short trigger prompt.

To add a ranking-aware signal, we derive differentiable item scores from the same forward pass. Let $\mathcal{T}_i$ denote the token positions of the $i$-th score line in $y^\star$. We define the model confidence for item $i$ as
\begin{equation}
\hat{s}_i
=
\frac{1}{|\mathcal{T}_i|}
\sum_{t\in\mathcal{T}_i}
\log
P_{\theta+\Delta W(\phi)}
\left(
y_t^\star
\mid
y_{<t}^\star,c_{\mathrm{trig}}
\right).
\label{eq:s_hat}
\end{equation}
This score remains differentiable with respect to $\phi$ and serves as a proxy for how well the model supports the teacher-provided score line.

We then apply a LambdaRank-style NDCG loss over item pairs. Let $g_i=2^{s_i}-1$ be the gain, $r_i$ the teacher rank, $\delta_i=1/\log_2(r_i+2)$ the discount, and $\mathrm{IDCG}=\sum_i g_i\delta_i$. The NDCG change from swapping items $i$ and $j$ is
\begin{equation}
|\Delta\mathrm{NDCG}_{ij}|
=
\frac{|g_i-g_j|\,|\delta_i-\delta_j|}{\mathrm{IDCG}}.
\label{eq:dndcg}
\end{equation}
The ranking loss is
\begin{equation}
\mathcal{L}_{\mathrm{rank}}(\phi)
=
\frac{1}{Z}
\sum_{\substack{i,j:\\ s_i>s_j}}
|\Delta\mathrm{NDCG}_{ij}|
\,
\mathrm{softplus}
\left(
-(\hat{s}_i-\hat{s}_j)
\right),
\label{eq:l_rank}
\end{equation}
where $Z=|\{(i,j):s_i>s_j\}|$. This term emphasizes pairs whose mis-ordering would have a larger impact on NDCG.

The final training objective is
\begin{equation}
\mathcal{L}_{\mathrm{total}}(\phi)
=
\mathcal{L}_{\mathrm{NLL}}(\phi)
+
\lambda
\mathcal{L}_{\mathrm{rank}}(\phi),
\label{eq:l_total}
\end{equation}
where $\lambda$ balances token-level imitation and listwise ordering. Since $\theta$ is frozen, gradients flow only through the synthesized adapter $\Delta W(\phi)$ into the hypernetwork $\mathcal{H}_\phi$.

\section{Experiments}
\label{sec:experiments}

To comprehensively evaluate the effectiveness of PRISMR, we conduct extensive experiments on a large-scale multimodal ranking benchmark. We aim to answer four primary research questions: \textbf{RQ1:} Can PRISMR eradicate generative parse collapse in listwise ranking and improve ranking quality, and how does it compare with prompt-engineering and format-enforcement baselines such as RankGPT, LLMLingua-2, and constrained decoding? \textbf{RQ2:} How well does PRISMR generalize beyond the training-time list length, and what are its practical inference-efficiency benefits? \textbf{RQ3:} Does prompt internalization remain effective in the pointwise setting, including when applied on top of a fine-tuned backbone? \textbf{RQ4:} Does PRISMR transfer across domains without additional training, and what does this reveal about the source of parse collapse?

\subsection{Experimental Setup}
\label{subsec:setup}
\textbf{Dataset.}
We construct a domain-specific multimodal review-ranking benchmark from Amazon Reviews 2023~\citep{hou2024bridging}. As summarized in Table~\ref{tab:dataset_stats}, each example contains 10--100 user reviews associated with the same product, where each review may include a title, textual content, and user-uploaded images. The task is to produce a listwise ranking of the reviews according to multiple quality dimensions, including multimodal content quality, semantic relevance, image quality, visual appeal, and expected helpfulness to users.
To validate that the induced benchmark rankings are aligned with human preferences, we conduct a pairwise preference evaluation. Because full listwise review ranking is inherently subjective and can exhibit low inter-annotator agreement, we sample 3,434 review pairs from the benchmark and collect pairwise judgments from both human annotators and an LLM judge. Specifically, we use Gemini-2.5-Pro as the LLM judge, which achieves 99.3\% agreement with the benchmark pairwise ordering. In parallel, two human annotators are asked to choose the preferred review in each pair according to the same quality criteria. The benchmark ordering achieves 90.6\% agreement with the aggregated human preferences. Notably, the agreement between the two annotators is also around 90\% (88.7\%), reflecting the inherent subjectivity of review ranking. These results suggest that our benchmark is well aligned with human preferences.


\begin{table}[h]
\centering
\small
\caption{Statistics of the constructed multimodal review benchmark.}
\label{tab:dataset_stats}
\begin{tabular}{l|rrr|rrrr}
\toprule
\textbf{Category} & \textbf{Products} & \textbf{Reviews} & \textbf{AvgRev} & \textbf{AvgLen} & \textbf{AvgScore} & \textbf{AvgImg} & \textbf{\%ImgRev} \\
\midrule
Baby Products  &  1,815 &  169,237 &  93.2 & 42.5 & 6.57 & 0.31 & 19.1\% \\
Amazon Fashion &  3,140 &  245,206 &  78.1 & 26.5 & 6.21 & 0.11 &  7.1\% \\
Software       &  5,007 &  466,111 &  93.1 & 33.6 & 5.89 & 0.01 &  1.0\% \\
\midrule
\textbf{Total / Avg.} & \textbf{9,962} & \textbf{880,554} & \textbf{88.4} & \textbf{33.3} & \textbf{6.11} & \textbf{0.10} & \textbf{6.2\%} \\
\bottomrule
\end{tabular}
\end{table}

\textbf{Baselines and Metrics.} We use Qwen3-VL-8B~\citep{bai2025qwen3} as the shared backbone. We compare PRISMR (default: $\alpha$-mode for $N\!\le\!50$, $\beta$-mode for $N\!>\!50$) with Base (zero-shot score-by-score prompting), RankGPT~\citep{sun2023chatgpt} (direct permutation generation), LLMLingua~\citep{jiang2023llmlingua,pan2024llmlingua}, Constrained-decoding Base (iterative decoding with one enforced line per candidate), and SFT Base. To analyse the contribution of each synthesis mode in isolation, we additionally report PRISMR ($\beta$-mode, mean pooling) at all $N$ as an ablation of the same checkpoint. We report NDCG@K for $K \in \{1,3,5,10\}$.

\textbf{Training hyperparameters.}
We use Qwen3-VL-8B as the frozen base model and train only the global hypernetwork $\mathcal{H}_\phi$ with 5.8$\times$10$^{8}$ parameters. $\mathcal{H}_\phi$ is a 6-layer Perceiver-style cross-attention encoder with 128 latent tokens of width 1024, 8 heads, GeLU MLPs with expansion ratio 4, RMSNorm, and two zero-initialized heads that output LoRA factors for each targeted projection. We use rank $r=2$ adapters on the $\{q,k,v,o\}$ projections of all 36 transformer blocks.
We train for 3 epochs using AdamW, learning rate $2\times10^{-5}$, $\beta_1=0.9$, $\beta_2=0.999$, no weight decay, and $\lambda=0.5$ in Eq.~\ref{eq:l_total}. The batch size is 1 per GPU with gradient accumulation 8. We use bfloat16 autocast, flash-attention 2, and gradient checkpointing on the frozen base. 
The data split is obtained by hashing product\_id with validation ratio 0.1 and seed 42. Multi-seed results for seeds $\{7,17\}$ are given in Appendix~\ref{app:training}. Training takes about 5 hours on one NVIDIA B200 or 2.5 hours on two B200 GPUs with DDP.

\paragraph{Parse rate.} For a list of size $N$, slot $i$ is counted as parsed if a fixed regex extracts a valid score for index $i \in [1,N]$ from the model output. We ignore duplicate indices after the first match and clip scores to $[1,10]$. We then define
\[
\mathrm{ParseRate}
=
\frac{1}{PN}
\sum_{p=1}^{P}\sum_{i=1}^{N}
\mathbb{1}\!\left[(p,i)\ \text{is parsed}\right].
\]
For NDCG computation, unparsed slots are assigned a default score of 5.0, which favors failed methods by providing a neutral fallback rather than the worst rank. A detailed failure-mode breakdown is given in Appendix~\ref{app:failure_decomp}.

\subsection{Eradicating Parse Collapse in Listwise Ranking (RQ1)}
\label{subsec:exp_listwise}

Table~\ref{tab:listwise_base} evaluates listwise ranking on multimodal \textit{Baby\_Products} with one image per review and $N \in \{10,20,50,100\}$.
Qwen3-VL exhibits severe parse collapse: parse rate drops from 11.3\% at $N{=}10$ to 2.0\% at $N{=}50$ and 1.0\% at $N{=}100$, with failures dominated by silent omissions rather than malformed outputs (Appendix~\ref{app:failure_decomp}). Prompt engineering only partially addresses this problem. RankGPT maintains high parse rate but its NDCG@10 still degrades with list length, while LLMLingua provides little benefit. Constrained decoding guarantees 100\% parse rate, yet remains substantially below PRISMR in NDCG@10, especially at larger $N$, showing that output-format enforcement alone fixes parsing but not long-context ranking quality.

PRISMR, run in its default configuration ($\alpha$-mode for $N\!\le\!50$, $\beta$-mode for $N\!>\!50$), achieves near-perfect parse rate and the best NDCG@10 at $N{=}10,20,50,100$. Its margin over the strongest prompt-based baseline widens as $N$ grows, indicating that the gain is not merely better format compliance but stronger listwise ranking under long multimodal context. 
To isolate the effect of the synthesis mode itself, we also report each mode in isolation across all $N$: $\beta$-mode alone already recovers most of the gain and maintains perfect parse rate, suggesting that the main benefit comes from hypernetwork-based parametric internalization, while $\alpha$-mode adds a smaller but consistent in-regime improvement. 

\begin{table}[ht]
\centering
\caption{Listwise ranking on Qwen3-VL-8B (\textit{Baby\_Products}, 1 image/review, 30 validation products per cell). PRISMR outperforms prompt-based baselines at every $N$. The bottom block reports the two synthesis modes of the same trained PRISMR checkpoint.} 
\label{tab:listwise_base}
\resizebox{\columnwidth}{!}{%
\begin{tabular}{l|cccc|cccc}
\toprule
 & \multicolumn{4}{c|}{\textbf{Parse Rate} (\%)} & \multicolumn{4}{c}{\textbf{NDCG@10}} \\
\textbf{Method} & $N{=}10$ & $N{=}20$ & $N{=}50$ & $N{=}100$\textsuperscript{$\ddagger$} & $N{=}10$ & $N{=}20$ & $N{=}50$ & $N{=}100$\textsuperscript{$\ddagger$} \\
\midrule
\multicolumn{9}{l}{\emph{Prompt-engineering / format-enforcement, no extra training:}} \\
Qwen3-VL & 11.3 & 8.2 & 2.0 & 1.0 & 0.767 & 0.553 & 0.418 & 0.341 \\
RankGPT \citep{sun2023chatgpt} & 90.0 & 93.8 & 92.6 & 91.0 & 0.799 & 0.668 & 0.597 & 0.538 \\
LLMLingua \citep{pan2024llmlingua} & 10.0 & 5.0 & 2.0 & -- & 0.734 & 0.513 & 0.399 & -- \\
Constrained-decoding Base & 100.0 & 100.0 & 100.0 & 100.0 & 0.910 & 0.858 & 0.789 & 0.757 \\
\midrule
\multicolumn{9}{l}{\emph{Trained on the same teacher data (no hypernet):}} \\
Standard SFT (LoRA $r{=}16$)\textsuperscript{$\ast$} & 67.0 & 39.3 & 37.9 & -- & 0.847 & 0.651 & 0.587 & -- \\
Constrained-decoding\textsuperscript{$\ast$} & 100.0 & 100.0 & 100.0 & -- & 0.963 & 0.947 & 0.916 & -- \\
RankGPT-SFT\textsuperscript{$\ast$} & 88.7 & 97.5 & 98.7 & -- & 0.886 & 0.696 & 0.780 & -- \\
\midrule
\multicolumn{9}{l}{\emph{PRISMR (single trained checkpoint, two synthesis modes; ours):}} \\
PRISMR ($\beta$-mode) \textsuperscript{$\dagger$} & 100.0 & 100.0 & 100.0 & \textbf{100.0} & 0.956 & 0.925 & 0.906 & \textbf{0.882} \\
\rowcolor{gray!10} PRISMR ($\alpha$-mode)\textsuperscript{$\flat$} & \textbf{100.0} & \textbf{100.0} & \textbf{99.9} & 85.1 & \textbf{0.974} & \textbf{0.958} & \textbf{0.947} & 0.335 \\
\bottomrule
\end{tabular}}
\\[3pt]\footnotesize{
$\ast$ Fine-tuning baselines use the same teacher labels and listwise prompts as PRISMR. \\
$\dagger$ $\beta$-mode: mean-pool synthesis of the same checkpoint (same hypernetwork as $\alpha$-mode). \\
$\flat$ Default PRISMR rule: use $\alpha$-mode for $N\!\le\!50$ (shaded; in-regime), and $\beta$-mode for $N\!>\!50$ (extrapolation). The two rows therefore correspond to one method with a length-adaptive synthesis switch. \\
$\ddagger$ $N{=}100$ is evaluated only for extrapolation.}
\end{table}

\subsection{Length Generalization and Inference Efficiency (RQ2)}
\label{subsec:exp_efficiency}
\begin{figure}[ht]
\centering
\includegraphics[width=\columnwidth]{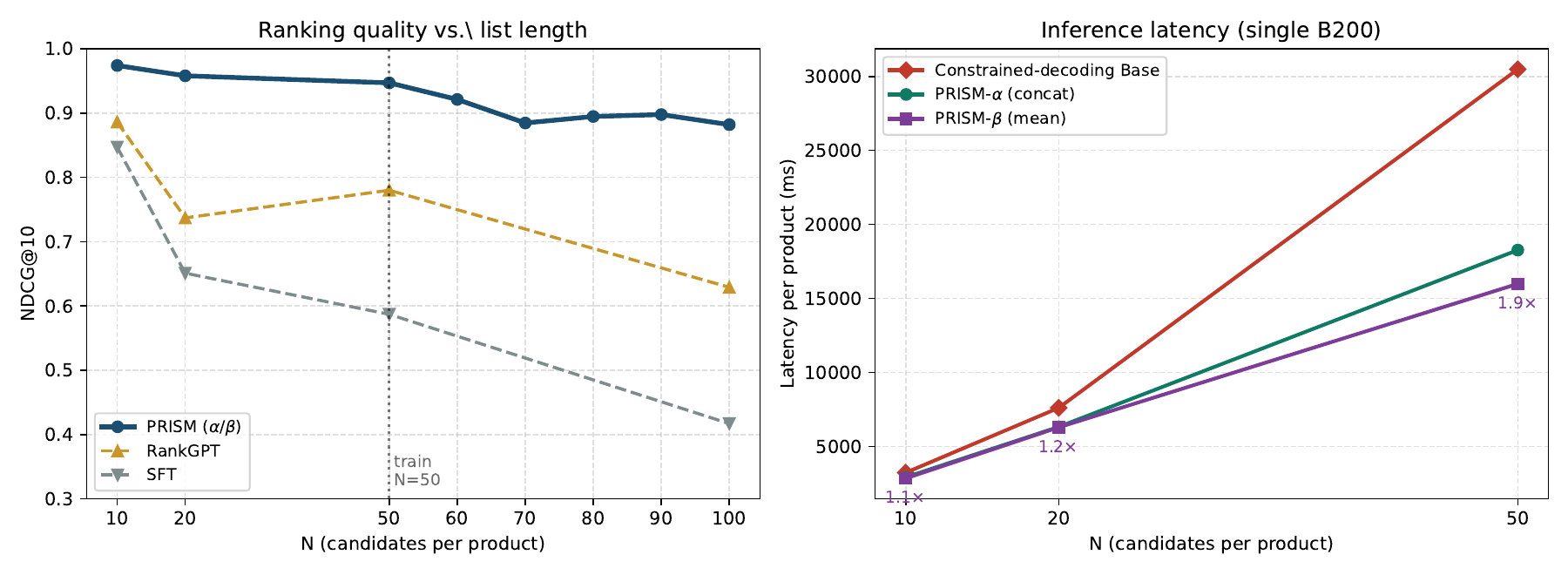}
\caption{\textbf{Left:} NDCG@10 vs.\ list length $N$. Both PRISMR modes are produced by the same checkpoint trained at $N{=}50$ (dotted line). $\alpha$-mode performs best in range ($N\!\le\!50$), while $\beta$-mode is more robust for extrapolation ($N\!>\!50$), motivating the default switch at $N{=}50$. The resulting PRISMR envelope outperforms the trained baselines across all $N$. \textbf{Right:} per-product latency on a single B200 (batch size 1, mean over 5 products). PRISMR is consistently faster than constrained-decoding Base, with speed-up increasing from $1.1\times$ at $N{=}10$ to $1.7\times$ at $N{=}50$.}

\label{fig:crossover}
\end{figure}

We examine how PRISMR's two synthesis modes behave as the candidate list grows beyond the training-time regime, and use the resulting curves to justify the $N\!=\!50$ mode-switch threshold in PRISMR's default rule. Figure~\ref{fig:crossover} (left) plots NDCG@10 as a function of list length $N$; the underlying hypernetwork is trained only up to $N{=}50$. In the in-range regime ($N\!\le\!50$), $\alpha$-mode consistently achieves the strongest ranking quality and remains clearly above the trained non-hypernetwork baselines, including Standard SFT and RankGPT-SFT. Beyond the training length, the two modes begin to diverge: $\alpha$-mode degrades sharply once the effective rank-dependent adapter width grows far beyond what was seen during training, whereas $\beta$-mode remains substantially more stable and overtakes $\alpha$-mode at larger $N$. The crossover sits near the training-time list length and motivates PRISMR's default length-adaptive synthesis rule (\textbf{$\alpha$-mode for $N\!\le\!50$; $\beta$-mode for $N\!>\!50$}), so that a single trained PRISMR checkpoint stays above the trained baselines across all evaluated list lengths.

Figure~\ref{fig:crossover} (right) shows that this improvement in ranking quality is accompanied by favorable inference cost. Measured by end-to-end wall-clock latency per product on a single NVIDIA B200 with batch size 1, PRISMR is consistently faster than constrained-decoding Base, and the gap widens as $N$ increases.
The reason is structural. Constrained decoding still performs generation over the full long context and therefore repeatedly incurs the cost of long-context attention, whereas PRISMR amortizes review conditioning into a one-time hypernetwork pass and then ranks using a compact parameterized representation. As the candidate list becomes longer, avoiding repeated long-context decoding becomes increasingly beneficial.
\subsection{Pointwise Internalization (RQ3)}
\label{subsec:exp_pointwise}

Beyond listwise ranking, we evaluate PRISMR in the pointwise setting, where the same long prompt must be re-encoded for each candidate, creating substantial redundancy. As shown in Table~\ref{tab:pointwise}, prompt internalization improves ranking quality across both text-only and multimodal inputs. On the unfine-tuned backbone, PRISMR raises NDCG@1 from 0.7878 to 0.9589 in the text-only setting and from 0.8302 to 0.9585 in the multimodal setting. The gains persist after finetuning, with SFT + PRISMR reaching 0.9848 NDCG@1 versus 0.9566 for SFT Base. These results show that parametric internalization also transfers to pointwise ranking and remains effective for multimodal inputs.


\begin{table}[ht]
\centering
\caption{Pointwise ranking performance under different prompt-internalization strategies and modalities. Here, \textit{sys} denotes internalizing only the system prompt into the LoRA, while \textit{all} denotes internalizing the full prompt.}
\label{tab:pointwise}
\small
\begin{tabular}{llcccc}
\toprule
\textbf{Setting} & \textbf{Method} & \textbf{NDCG@1} & \textbf{NDCG@3} & \textbf{NDCG@5} & \textbf{NDCG@10} \\
\midrule

\multicolumn{6}{c}{\textit{Base model (un-finetuned)}} \\
\midrule
Text-only  & Base (sys)         & 0.7878 & 0.7997 & 0.8572 & 0.9114 \\
Text-only  & PRISMR (sys)        & 0.9552 & 0.9700 & 0.9795 & 0.9858 \\
Text-only  & PRISMR (all)        & 0.9589 & 0.9684 & 0.9744 & 0.9850 \\
\midrule
Multimodal & Base (all)         & 0.8302 & 0.8878 & 0.9081 & 0.9402 \\
Multimodal & PRISMR (sys)        & 0.9585 & \textbf{0.9775} & 0.9827 & 0.9881 \\
Multimodal & PRISMR (all)        & 0.9517 & 0.9628 & 0.9747 & 0.9823 \\
\midrule

\multicolumn{6}{c}{\textit{SFT model (finetuned)}} \\
\midrule
Multimodal & SFT Base (all)     & 0.9566 & 0.9547 & 0.9768 & 0.9823 \\
Multimodal & SFT + PRISMR (all)  & \textbf{0.9848} & 0.9725 & \textbf{0.9831} & \textbf{0.9890} \\
\bottomrule
\end{tabular}
\end{table}


\subsection{Cross-Domain Transfer (RQ4)}
\label{subsec:exp_xdomain}
We next test whether PRISMR's gains are domain-specific or transfer beyond the training category. To this end, we evaluate the checkpoint trained on \textit{Baby\_Products} directly on the disjoint \textit{Amazon\_Fashion} category, without any additional training. As shown in Table~\ref{tab:xdomain}, PRISMR ($\alpha$-mode) transfers cleanly across domains: it retains essentially perfect parse rate at all list lengths and achieves strong NDCG@10 on Fashion (0.982, 0.976, and 0.962 at $N{=}10,20,50$), remaining close to the in-domain results on Baby\_Products. By contrast, the Base score-by-score model exhibits the same parse-collapse pattern on the new domain, with parse rate dropping to 14.0\%, 5.0\%, and 2.0\% as $N$ increases. This shows that parse collapse is primarily a property of the backbone-and-decoding protocol rather than of any particular domain. Although \textit{Amazon\_Fashion} has a somewhat different score distribution and appears slightly easier for both methods, the absolute gap between PRISMR and the Base model remains large, exceeding 0.4 NDCG@10 at every $N$.

\begin{table}[ht]
\centering
\caption{Zero-shot cross-domain transfer from \textit{Baby\_Products} to \textit{Amazon\_Fashion}. The model is trained on \textit{Baby\_Products} and evaluated on \textit{Amazon\_Fashion} without additional training.}
\label{tab:xdomain}
\small
\begin{tabular}{l|ccc|ccc}
\toprule
 & \multicolumn{3}{c|}{\textbf{Parse Rate} (\%)} & \multicolumn{3}{c}{\textbf{NDCG@10}} \\
\textbf{Method} & $N{=}10$ & $N{=}20$ & $N{=}50$ & $N{=}10$ & $N{=}20$ & $N{=}50$ \\
\midrule
Base (score-by-score) & 14.0 & 5.0 & 2.0 & 0.806 & 0.525 & 0.382 \\
\rowcolor{gray!10} \textbf{PRISMR ($\alpha$-mode) (Baby$\to$Fashion, ours)} & \textbf{100.0} & \textbf{100.0} & \textbf{99.9} & \textbf{0.982} & \textbf{0.976} & \textbf{0.962} \\
\bottomrule
\end{tabular}
\end{table}

\section{Limitations}
\label{sec:limitations}

PRISMR has several limitations. First, its format compliance is empirical rather than formally guaranteed. Although it achieves a $\ge 99.9\%$ parse rate in the trained regime, the $\alpha$-mode synthesis can degrade far beyond the training list length; PRISMR therefore switches to its $\beta$-mode at $N\!>\!50$ for out-of-regime deployment, which is part of the default rule rather than a separate method.
Second, review ranking is subjective, and no single ordering can perfectly reflect all user preferences, especially when candidate reviews are close in quality. 


\section{Conclusion}
\label{sec:conclusion}

We identify parse collapse as a key failure mode of generative listwise ranking: for long multimodal inputs, the base vision-language model often terminates early and leaves much of the list unparsed. Prompt compression and alternative prompting formats do not resolve this issue, suggesting that the bottleneck lies in the fragile long-context generative output protocol.
We propose PRISMR, a hypernetwork-based conditioning method that encodes each candidate review as a per-item LoRA and composes them into an instance-specific ranking module. PRISMR nearly eliminates parse collapse, improves over prompting and constrained-decoding baselines, remains effective for pointwise ranking, and transfers across domains. We further observe a synthesis-mode trade-off: $\alpha$-mode is stronger in range, whereas $\beta$-mode is more robust under length extrapolation; thus PRISMR uses $\alpha$ for $N\!\le\!50$ and $\beta$ for $N\!>\!50$.
Overall, PRISMR shows that moving per-candidate information from transient prompt tokens into structured parameter space offers a practical alternative to increasingly complex prompting or decoding heuristics.

% ...

\bibliographystyle{plainnat}
\bibliography{references}

\begin{thebibliography}{34}
\providecommand{\natexlab}[1]{#1}
\providecommand{\url}[1]{\texttt{#1}}
\expandafter\ifx\csname urlstyle\endcsname\relax
  \providecommand{\doi}[1]{doi: #1}\else
  \providecommand{\doi}{doi: \begingroup \urlstyle{rm}\Url}\fi

\bibitem[Achiam et~al.(2023)Achiam, Adler, Agarwal, Ahmad, Akkaya, Aleman,
  Almeida, Altenschmidt, Altman, Anadkat, et~al.]{achiam2023gpt}
Josh Achiam, Steven Adler, Sandhini Agarwal, Lama Ahmad, Ilge Akkaya,
  Florencia~Leoni Aleman, Diogo Almeida, Janko Altenschmidt, Sam Altman,
  Shyamal Anadkat, et~al.
\newblock Gpt-4 technical report.
\newblock \emph{arXiv preprint arXiv:2303.08774}, 2023.

\bibitem[Bai et~al.(2025)Bai, Cai, Chen, Chen, Chen, Cheng, Deng, Ding, Gao,
  Ge, et~al.]{bai2025qwen3}
Shuai Bai, Yuxuan Cai, Ruizhe Chen, Keqin Chen, Xionghui Chen, Zesen Cheng,
  Lianghao Deng, Wei Ding, Chang Gao, Chunjiang Ge, et~al.
\newblock Qwen3-vl technical report.
\newblock \emph{arXiv preprint arXiv:2511.21631}, 2025.

\bibitem[Cai et~al.(2025)Cai, Gao, Zhang, Shi, Zhang, Bao, Wang, and
  Feng]{cai2025k}
Shihao Cai, Chongming Gao, Yang Zhang, Wentao Shi, Jizhi Zhang, Keqin Bao,
  Qifan Wang, and Fuli Feng.
\newblock K-order ranking preference optimization for large language models.
\newblock In \emph{Findings of the Association for Computational Linguistics:
  ACL 2025}, pages 4844--4859, 2025.

\bibitem[Charakorn et~al.(2025)Charakorn, Cetin, Tang, and
  Lange]{charakorn2025text}
Rujikorn Charakorn, Edoardo Cetin, Yujin Tang, and Robert~Tjarko Lange.
\newblock Text-to-lora: Instant transformer adaption.
\newblock \emph{arXiv preprint arXiv:2506.06105}, 2025.

\bibitem[Charakorn et~al.(2026)Charakorn, Cetin, Uesaka, and
  Lange]{charakorn2026doc}
Rujikorn Charakorn, Edoardo Cetin, Shinnosuke Uesaka, and Robert~Tjarko Lange.
\newblock Doc-to-lora: Learning to instantly internalize contexts.
\newblock \emph{arXiv preprint arXiv:2602.15902}, 2026.

\bibitem[Dong et~al.(2025)Dong, Chang, Deik, Li, Tang, and
  Liu]{dong2025mmdocir}
Kuicai Dong, Yujing Chang, Derrick Goh~Xin Deik, Dexun Li, Ruiming Tang, and
  Yong Liu.
\newblock Mmdocir: Benchmarking multimodal retrieval for long documents.
\newblock In \emph{Proceedings of the 2025 Conference on Empirical Methods in
  Natural Language Processing}, pages 30959--30993, 2025.

\bibitem[Gera et~al.(2025)Gera, Boni, Perlitz, Bar-Haim, Eden, and
  Yehudai]{gera2025justrank}
Ariel Gera, Odellia Boni, Yotam Perlitz, Roy Bar-Haim, Lilach Eden, and Asaf
  Yehudai.
\newblock Justrank: Benchmarking llm judges for system ranking.
\newblock In \emph{Proceedings of the 63rd Annual Meeting of the Association
  for Computational Linguistics (Volume 1: Long Papers)}, pages 682--712, 2025.

\bibitem[Gupta et~al.(2025)Gupta, You, Bhojanapalli, Kumar, Dhillon, and
  Yu]{gupta2025scalable}
Nilesh Gupta, Chong You, Srinadh Bhojanapalli, Sanjiv Kumar, Inderjit Dhillon,
  and Felix Yu.
\newblock Scalable in-context ranking with generative models.
\newblock \emph{arXiv preprint arXiv:2510.05396}, 2025.

\bibitem[Hou et~al.(2024)Hou, Li, He, Yan, Chen, and McAuley]{hou2024bridging}
Yupeng Hou, Jiacheng Li, Zhankui He, An~Yan, Xiusi Chen, and Julian McAuley.
\newblock Bridging language and items for retrieval and recommendation.
\newblock \emph{arXiv preprint arXiv:2403.03952}, 2024.

\bibitem[Hu et~al.(2022)Hu, Shen, Wallis, Allen-Zhu, Li, Wang, Wang, Chen,
  et~al.]{hu2022lora}
Edward~J Hu, Yelong Shen, Phillip Wallis, Zeyuan Allen-Zhu, Yuanzhi Li, Shean
  Wang, Liang Wang, Weizhu Chen, et~al.
\newblock Lora: Low-rank adaptation of large language models.
\newblock \emph{Iclr}, 1\penalty0 (2):\penalty0 3, 2022.

\bibitem[Ivison et~al.(2023)Ivison, Bhagia, Wang, Hajishirzi, and
  Peters]{ivison2023hint}
Hamish Ivison, Akshita Bhagia, Yizhong Wang, Hannaneh Hajishirzi, and Matthew~E
  Peters.
\newblock Hint: Hypernetwork instruction tuning for efficient zero- and
  few-shot generalisation.
\newblock \emph{Proceedings of the 61st Annual Meeting of the Association for
  Computational Linguistics (ACL)}, 2023.

\bibitem[Jiang et~al.(2026)Jiang, Yang, Wang, Zhang, and Lin]{jiang2026rlpo}
Hao Jiang, Zhi Yang, Annan Wang, Yichi Zhang, and Weisi Lin.
\newblock Rlpo: Residual listwise preference optimization for long-context
  review ranking.
\newblock \emph{arXiv preprint arXiv:2601.07449}, 2026.

\bibitem[Jiang et~al.(2023)Jiang, Wu, Lin, Yang, and Qiu]{jiang2023llmlingua}
Huiqiang Jiang, Qianhui Wu, Chin-Yew Lin, Yuqing Yang, and Lili Qiu.
\newblock Llmlingua: Compressing prompts for accelerated inference of large
  language models.
\newblock In \emph{Proceedings of the 2023 conference on empirical methods in
  natural language processing}, pages 13358--13376, 2023.

\bibitem[Liu et~al.(2024)Liu, Lin, Hewitt, Paranjape, Bevilacqua, Petroni, and
  Liang]{liu2024lost}
Nelson~F Liu, Kevin Lin, John Hewitt, Ashwin Paranjape, Michele Bevilacqua,
  Fabio Petroni, and Percy Liang.
\newblock Lost in the middle: How language models use long contexts.
\newblock \emph{Transactions of the association for computational linguistics},
  12:\penalty0 157--173, 2024.

\bibitem[Liu et~al.(2025{\natexlab{a}})Liu, Duan, Chen, Lu, Sun, and
  Mao]{liu2025llm4ranking}
Qi~Liu, Haozhe Duan, Yiqun Chen, Quanfeng Lu, Weiwei Sun, and Jiaxin Mao.
\newblock Llm4ranking: An easy-to-use framework of utilizing large language
  models for document reranking.
\newblock \emph{arXiv preprint arXiv:2504.07439}, 2025{\natexlab{a}}.

\bibitem[Liu et~al.(2025{\natexlab{b}})Liu, Qin, Wu, Shen, Khalman, Joshi,
  Zhao, Saleh, Baumgartner, Liu, et~al.]{liu2025lipo}
Tianqi Liu, Zhen Qin, Junru Wu, Jiaming Shen, Misha Khalman, Rishabh Joshi, Yao
  Zhao, Mohammad Saleh, Simon Baumgartner, Jialu Liu, et~al.
\newblock Lipo: Listwise preference optimization through learning-to-rank.
\newblock In \emph{Proceedings of the 2025 Conference of the Nations of the
  Americas Chapter of the Association for Computational Linguistics: Human
  Language Technologies (Volume 1: Long Papers)}, pages 2404--2420,
  2025{\natexlab{b}}.

\bibitem[Liu et~al.(2025{\natexlab{c}})Liu, Ma, Zhu, Su, Wang, Yin, and
  Dou]{liu2025coranking}
Wenhan Liu, Xinyu Ma, Yutao Zhu, Lixin Su, Shuaiqiang Wang, Dawei Yin, and
  Zhicheng Dou.
\newblock Coranking: Collaborative ranking with small and large ranking agents.
\newblock \emph{arXiv preprint arXiv:2503.23427}, 2025{\natexlab{c}}.

\bibitem[Mu et~al.(2023)Mu, Li, and Goodman]{mu2023gisting}
Jesse Mu, Xiang~Lisa Li, and Noah~D Goodman.
\newblock Learning to compress prompts with gist tokens.
\newblock In \emph{Advances in Neural Information Processing Systems
  (NeurIPS)}, 2023.

\bibitem[Pan et~al.(2024)Pan, Wu, Jiang, Xia, Luo, Zhang, Lin, R{\"u}hle, Yang,
  Lin, et~al.]{pan2024llmlingua}
Zhuoshi Pan, Qianhui Wu, Huiqiang Jiang, Menglin Xia, Xufang Luo, Jue Zhang,
  Qingwei Lin, Victor R{\"u}hle, Yuqing Yang, Chin-Yew Lin, et~al.
\newblock Llmlingua-2: Data distillation for efficient and faithful
  task-agnostic prompt compression.
\newblock In \emph{Findings of the Association for Computational Linguistics:
  ACL 2024}, pages 963--981, 2024.

\bibitem[Phang et~al.(2023)Phang, Mao, He, and Chen]{phang2023hypertuning}
Jason Phang, Yi~Mao, Pengcheng He, and Weizhu Chen.
\newblock Hypertuning: Toward adapting large language models without
  back-propagation.
\newblock \emph{Proceedings of the 40th International Conference on Machine
  Learning (ICML)}, 2023.

\bibitem[Pradeep et~al.(2023)Pradeep, Sharifymoghaddam, and
  Lin]{pradeep2023rankzephyr}
Ronak Pradeep, Sahel Sharifymoghaddam, and Jimmy Lin.
\newblock Rankzephyr: Effective and robust zero-shot listwise reranking is a
  breeze!
\newblock \emph{arXiv preprint arXiv:2312.02724}, 2023.

\bibitem[Qiu et~al.(2022)Qiu, Hu, Zhong, Zhang, and Yang]{qiu2022large}
Zi-Hao Qiu, Quanqi Hu, Yongjian Zhong, Lijun Zhang, and Tianbao Yang.
\newblock Large-scale stochastic optimization of ndcg surrogates for deep
  learning with provable convergence.
\newblock \emph{arXiv preprint arXiv:2202.12183}, 2022.

\bibitem[Rafailov et~al.(2023)Rafailov, Sharma, Mitchell, Manning, Ermon, and
  Finn]{rafailov2023direct}
Rafael Rafailov, Archit Sharma, Eric Mitchell, Christopher~D Manning, Stefano
  Ermon, and Chelsea Finn.
\newblock Direct preference optimization: Your language model is secretly a
  reward model.
\newblock \emph{Advances in neural information processing systems},
  36:\penalty0 53728--53741, 2023.

\bibitem[Reddy et~al.(2024)Reddy, Doo, Xu, Sultan, Swain, Sil, and
  Ji]{reddy2024first}
Revanth~Gangi Reddy, JaeHyeok Doo, Yifei Xu, Md~Arafat Sultan, Deevya Swain,
  Avirup Sil, and Heng Ji.
\newblock First: Faster improved listwise reranking with single token decoding.
\newblock In \emph{Proceedings of the 2024 Conference on Empirical Methods in
  Natural Language Processing}, pages 8642--8652, 2024.

\bibitem[Ren et~al.(2025)Ren, Wang, Zhou, Zhao, Wang, Liu, Wen, and
  Chua]{ren2025self}
Ruiyang Ren, Yuhao Wang, Kun Zhou, Wayne~Xin Zhao, Wenjie Wang, Jing Liu,
  Ji-Rong Wen, and Tat-Seng Chua.
\newblock Self-calibrated listwise reranking with large language models.
\newblock In \emph{Proceedings of the ACM on Web Conference 2025}, pages
  3692--3701, 2025.

\bibitem[Shao et~al.(2024)Shao, Wang, Zhu, Xu, Song, Bi, Zhang, Zhang, Li, Wu,
  et~al.]{shao2024deepseekmath}
Zhihong Shao, Peiyi Wang, Qihao Zhu, Runxin Xu, Junxiao Song, Xiao Bi, Haowei
  Zhang, Mingchuan Zhang, YK~Li, Yang Wu, et~al.
\newblock Deepseekmath: Pushing the limits of mathematical reasoning in open
  language models.
\newblock \emph{arXiv preprint arXiv:2402.03300}, 2024.

\bibitem[Shin et~al.(2025)Shin, Ji, Gong, Kim, Choi, and
  Seo]{shin2025generative}
Haebin Shin, Lei Ji, Yeyun Gong, Sungdong Kim, Eunbi Choi, and Minjoon Seo.
\newblock Generative prompt internalization.
\newblock In \emph{Proceedings of the 2025 Conference of the Nations of the
  Americas Chapter of the Association for Computational Linguistics: Human
  Language Technologies (Volume 1: Long Papers)}, pages 7338--7363, 2025.

\bibitem[Snell et~al.(2022)Snell, Klein, and
  Zhong]{snell2022contextdistillation}
Charlie Snell, Dan Klein, and Ruiqi Zhong.
\newblock Learning by distilling context.
\newblock \emph{arXiv preprint arXiv:2209.15189}, 2022.

\bibitem[Sun et~al.(2023)Sun, Yan, Ma, Wang, Ren, Chen, Yin, and
  Ren]{sun2023chatgpt}
Weiwei Sun, Lingyong Yan, Xinyu Ma, Shuaiqiang Wang, Pengjie Ren, Zhumin Chen,
  Dawei Yin, and Zhaochun Ren.
\newblock Is chatgpt good at search? investigating large language models as
  re-ranking agents.
\newblock In \emph{Proceedings of the 2023 conference on empirical methods in
  natural language processing}, pages 14918--14937, 2023.

\bibitem[Team et~al.(2023)Team, Anil, Borgeaud, Alayrac, Yu, Soricut,
  Schalkwyk, Dai, Hauth, Millican, et~al.]{team2023gemini}
Gemini Team, Rohan Anil, Sebastian Borgeaud, Jean-Baptiste Alayrac, Jiahui Yu,
  Radu Soricut, Johan Schalkwyk, Andrew~M Dai, Anja Hauth, Katie Millican,
  et~al.
\newblock Gemini: a family of highly capable multimodal models.
\newblock \emph{arXiv preprint arXiv:2312.11805}, 2023.

\bibitem[Vendrow et~al.(2024)Vendrow, Pantazis, Shepard, Brostow, Jones,
  Mac~Aodha, Beery, and Van~Horn]{vendrow2024inquire}
Edward Vendrow, Omiros Pantazis, Alexander Shepard, Gabriel Brostow, Kate~E
  Jones, Oisin Mac~Aodha, Sara Beery, and Grant Van~Horn.
\newblock Inquire: A natural world text-to-image retrieval benchmark.
\newblock \emph{Advances in Neural Information Processing Systems},
  37:\penalty0 126500--126514, 2024.

\bibitem[Wu et~al.(2025)Wu, Surana, Xie, Shen, Xia, Yu, Rossi, Ammanabrolu, and
  McAuley]{wu2025context}
Junda Wu, Rohan Surana, Zhouhang Xie, Yiran Shen, Yu~Xia, Tong Yu, Ryan~A
  Rossi, Prithviraj Ammanabrolu, and Julian McAuley.
\newblock In-context ranking preference optimization.
\newblock \emph{arXiv preprint arXiv:2504.15477}, 2025.

\bibitem[Yang et~al.(2025)Yang, Li, Yang, Zhang, Hui, Zheng, Yu, Gao, Huang,
  Lv, et~al.]{yang2025qwen3}
An~Yang, Anfeng Li, Baosong Yang, Beichen Zhang, Binyuan Hui, Bo~Zheng, Bowen
  Yu, Chang Gao, Chengen Huang, Chenxu Lv, et~al.
\newblock Qwen3 technical report.
\newblock \emph{arXiv preprint arXiv:2505.09388}, 2025.

\bibitem[Zhao et~al.(2025)Zhao, Wang, and Yin]{zhao2025permutative}
Yang Zhao, Yixin Wang, and Mingzhang Yin.
\newblock Permutative preference alignment from listwise ranking of human
  judgments.
\newblock In \emph{Proceedings of the 2025 Conference on Empirical Methods in
  Natural Language Processing}, pages 310--334, 2025.

\end{thebibliography}

\appendix

\newpage
\section{Training Details}
\label{app:training}

\textbf{Hardware.}
Listwise PRISMR is trained on a single node with $2 \times$ NVIDIA B200 GPUs, each with 180\,GiB of memory. All evaluation-only experiments fit on a single B200 GPU.

\textbf{Optimizer and precision.}
We optimize the hypernetwork parameters with AdamW implemented in \texttt{torch.optim.AdamW}, using learning rate $2 \times 10^{-5}$, $\beta_1 = 0.9$, $\beta_2 = 0.999$, and no weight decay. Training uses mixed precision with \texttt{torch.autocast} in \texttt{bfloat16}. The base model is frozen throughout training. We enable gradient checkpointing on the frozen base model to reduce activation memory and keep both the base-model and hypernetwork activations resident during backpropagation.

\textbf{Schedule and batching.}
Unless otherwise stated, we train for three epochs. The per-GPU batch size is 1, with gradient accumulation over 8 steps, yielding an effective batch size of 16 across the two GPUs. For the Baby\_Products domain at $N=50$, this corresponds to approximately $1{,}633$ training products per epoch and three full passes over the training split.

\textbf{Hypernetwork architecture.}
The global hypernetwork $\mathcal{H}_\phi$ is a Perceiver-style cross-attention encoder followed by layer-specific linear LoRA heads. It produces candidate-conditioned LoRA increments for every targeted projection in all 36 transformer layers of Qwen3-VL-8B. For each candidate, the input context consists of the system prompt, product header, review text, and image patches. This context is tokenized by the base model tokenizer and vision encoder, and is consumed by a 6-layer cross-attention stack with 128 learnable latent tokens of width $d_{\mathrm{lat}}=1024$. Each layer uses 8 attention heads, GeLU MLPs with expansion factor 4, RMSNorm, and residual connections.

For each targeted projection $\ell$, two linear heads map the latent representation to LoRA factors
\[
A_i^{(\ell)} \in \mathbb{R}^{r \times d_{\mathrm{in}}}, 
\qquad
B_i^{(\ell)} \in \mathbb{R}^{d_{\mathrm{out}} \times r}.
\]
The $B$ heads are zero-initialized, so that each candidate-specific update
\[
\Delta W_i^{(\ell)} = B_i^{(\ell)} A_i^{(\ell)}
\]
is initialized to zero. We target the $\{q,k,v,o\}$ projections in every transformer block, resulting in $4 \times 36 = 144$ targeted projections. The per-candidate LoRA rank is $r=2$. The hypernetwork contains approximately $5.8 \times 10^8$ trainable parameters, while the frozen base model contains approximately $1.69 \times 10^{10}$ parameters. Under the $\alpha$-mode synthesis rule, the composite increment
\[
\Delta W_\alpha = \sum_i B_i A_i
\]
has effective rank at most $Nr$, as stated in Equation~\ref{eq:alpha_dw}.

\textbf{Vision settings.}
We use at most one image per review, with \texttt{max\_images\_per\_review = 1}. Images are resized subject to \texttt{image\_max\_pixels = $672 \times 672$ = 451\,584}. FlashAttention-2 is enabled for the base model.

\textbf{Loss and supervision.}
Training uses the token-level negative log-likelihood on teacher-curated score lines, defined in Equation~\ref{eq:l_nll}, together with the LambdaRank-NDCG auxiliary loss in Equation~\ref{eq:l_rank}. The auxiliary weight is $\lambda=0.5$. The differentiable per-item score readout $\hat{s}_i$ in Equation~\ref{eq:s_hat} partitions answer-span tokens by a cumulative sum over newline token ids. This allows the implementation to aggregate token-level quantities with a single \texttt{scatter\_add} operation. The pairwise LambdaRank term uses the closed-form $|\Delta \mathrm{NDCG}_{ij}|$ computed from teacher-induced ranks and discounts.

\textbf{Teacher model and filtering.}
The teacher model is GPT-5.4-2026-04-15, using a pinned snapshot. We filter teacher traces before training. A trace is discarded if the response: 
(i) omits any of the $N$ requested score lines; 
(ii) emits an out-of-range or repeated candidate index; or 
(iii) violates the required score-line regular expression. 
On Baby\_Products, this filtering procedure retains $79.4\%$ of teacher trajectories at $N=50$ and $94.1\%$ at $N=10$. The retained trajectories constitute $\mathcal{D}_{\mathrm{train}}$.


\textbf{Code, models, and licences.}
We will release the training pipeline, evaluation scripts, configuration files, and checkpoint hashes upon publication. The Amazon Reviews 2023 corpus is released under the Apache-2.0 licence; INQUIRE-Rerank under CC BY-NC 4.0; MMDocIR under MIT; and Qwen3-VL-8B under Apache-2.0.

\textbf{Data splits.}
We use a single-domain split with \texttt{val\_ratio = 0.1} and \texttt{data\_seed = 42}, yielding 1,633 training products and 182 validation products. Each cell in Table~\ref{tab:listwise_base} is computed over 30 validation products sampled under the same seed. The same validation products are used across methods, making comparisons paired.

\textbf{Multi-seed validation.}
To estimate seed-induced variance for the headline PRISMR results, we additionally train PRISMR in both $\alpha$-mode and $\beta$-mode at $N=50$ using two extra seeds, $\{7,17\}$, with otherwise identical hyperparameters. For each seed, we evaluate on a paired 30-product validation draw generated under the same seed. Table~\ref{tab:multiseed} reports the mean and standard deviation of NDCG@10 across these runs. The parse rate is $100\%$ for every evaluated cell. Across both PRISMR modes and all evaluated list lengths, the standard deviation does not exceed $0.0064$ NDCG@10, which is substantially smaller than the absolute gap between PRISMR and the strongest in-context baseline. For example, in Table~\ref{tab:listwise_base}, PRISMR $\alpha$-mode improves over RankGPT by $0.18$, $0.29$, and $0.37$ NDCG@10 at $N=10$, $N=20$, and $N=50$, respectively.

\begin{table}[ht]
\centering
\caption{Multi-seed NDCG@10 of PRISMR on Qwen3-VL-8B + Baby\_Products (mean $\pm$ std across two extra seeds $\{7, 17\}$, identical 30-val-product paired draw, parse rate $100\%$ throughout).}
\label{tab:multiseed}
\begin{tabular}{l|ccc}
\toprule
\textbf{Method} & $N{=}10$ & $N{=}20$ & $N{=}50$ \\ \midrule
PRISMR ($\alpha$-mode) (rank-dim concat) & $0.978 \pm 0.001$ & $0.962 \pm 0.006$ & $0.966 \pm 0.003$ \\
PRISMR ($\beta$-mode) (mean-pool)        & $0.981 \pm 0.005$ & $0.972 \pm 0.004$ & $0.958 \pm 0.003$ \\
\bottomrule
\end{tabular}
\end{table}

\section{Failure-mode decomposition}
\label{app:failure_decomp}

We classify every \emph{unparsed slot} produced by the score-by-score base into one of five mutually-overlapping categories: \texttt{missing\_count} (the model never emitted a line for that index), \texttt{hallucinated\_id} (an out-of-range index appeared in the output), \texttt{format\_error} (the output contained no parseable ``\texttt{i: <score>}'' substring at all), \texttt{length\_overflow} (the decoder exhausted \texttt{max\_new\_tokens} without finishing the list), and \texttt{repeated\_id} (an in-range index appeared more than once).

\begin{figure}[ht]
\centering
\includegraphics[width=0.65\columnwidth]{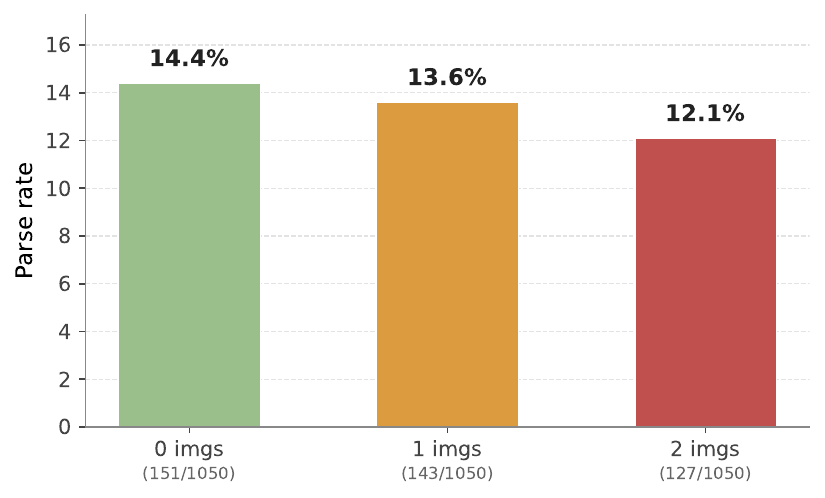}
\caption{Parse rate of the score-by-score Base model on Qwen3-VL-8B + Baby\_Products at $N=20$ under different image-density settings. Parse rate is low in all cases and decreases as more images are included, indicating that multimodal context exacerbates listwise parse collapse.}
\label{fig:fig3_failure_decomp}
\end{figure}

Empirically, the score-by-score Base model exhibits severe parse collapse at $N=20$ across all image-density settings. As shown in Figure~\ref{fig:fig3_failure_decomp}, the parse rate remains low even when no images are included: $14.4\%$ for text-only inputs, corresponding to $151/1050$ successfully parsed examples. Adding images further reduces the parse rate to $13.6\%$ with one image per review ($143/1050$) and $12.1\%$ with two images per review ($127/1050$). 

\section{Generalisation to Standard Multimodal IR Benchmarks}
\label{app:inquire}

To test whether PRISMR's structural fix transfers beyond the Amazon Reviews family, we run the same Qwen3-VL-8B + PRISMR checkpoint (trained on \textit{Baby\_Products} review-quality scoring) zero-shot on two off-the-shelf multimodal IR benchmarks: \textbf{INQUIRE-Rerank}~\citep{vendrow2024inquire}, an image-retrieval re-ranking benchmark constructed from iNaturalist 2024 with natural-language ecology queries (e.g.\ ``a mongoose standing upright alert''), and \textbf{MMDocIR}~\citep{dong2025mmdocir}, a multimodal document re-ranking benchmark over PDF page screenshots paired with VLM-extracted page text. For each of 30 randomly-drawn queries we form a listwise prompt with $N{=}10$ candidate items: for INQUIRE, 5 positives + 5 negatives sampled from the CLIP top-100 (binary relevance); for MMDocIR, the question's gold pages plus random non-gold pages from the same document.

\begin{table}[ht]
\centering
\caption{Generalisation to two off-the-shelf multimodal IR benchmarks at $N{=}10$ (30 queries each). The top block is zero-shot (Baby$\to$benchmark); the bottom block fine-tunes the same PRISMR/PRISMR ($\beta$-mode) checkpoint for two epochs on the benchmark's training split. Parse rate transfers cleanly in both regimes. In-list ranking quality does \emph{not} transfer zero-shot, but recovers substantially after a brief in-domain fine-tune; even so, fine-tuned PRISMR is statistically tied with the (training-free) constrained-decoding baseline on these short ($N{=}10$) image-only / document-page rerank tasks.}
\label{tab:inquire}
\begin{tabular}{l|cc|cc}
\toprule
 & \multicolumn{2}{c|}{\textbf{INQUIRE-Rerank}} & \multicolumn{2}{c}{\textbf{MMDocIR}} \\
\textbf{Method} & Parse \% & NDCG@10 & Parse \% & NDCG@10 \\ \midrule
\multicolumn{5}{l}{\emph{Zero-shot (Baby$\to$benchmark, no further training):}} \\
Base (score-by-score) & 10.0 & 0.707 & --- & --- \\
RankGPT \citep{sun2023chatgpt} & 90.0 & 0.728 & 90.0 & 0.761 \\
\textbf{Constrained-decoding Base} & \textbf{100.0} & \textbf{0.841} & \textbf{100.0} & \textbf{0.857} \\
PRISMR ($\alpha$-mode) (concat) & 100.0 & 0.705 & 100.0 & 0.753 \\ \midrule
\multicolumn{5}{l}{\emph{In-domain fine-tune (2 epochs on the benchmark training split):}} \\
PRISMR ($\beta$-mode) (mean) fine-tuned       & 100.0 & 0.831 & 100.0 & 0.834 \\
\rowcolor{gray!10} \textbf{PRISMR ($\alpha$-mode) (concat) fine-tuned} & \textbf{100.0} & 0.804 & \textbf{100.0} & \textbf{0.844} \\
\bottomrule
\end{tabular}
\end{table}

The result is consistent across both benchmarks and splits into two halves. Parse-rate fix transfers. PRISMR keeps a $100\,\%$ parse rate on tasks whose candidates (single nature photos for INQUIRE, document-page screenshots for MMDocIR) look nothing like its training distribution; the score-by-score Base collapses identically here as on Baby. In-list ranking-quality does not transfer zero-shot, but recovers under fine-tune. Zero-shot PRISMR's NDCG@10 of $0.705$ on INQUIRE matches the random baseline ($0.707$); inspecting outputs we observe position-dependent scores ($2.8, 2.9, 3.1, 3.0, \ldots$) rather than content-dependent ones, indicating that the hypernet---trained to extract text-quality features from review bodies---has no analogous signal to extract from a brief species caption or a document page screenshot. After two epochs of fine-tuning on the benchmark's own train split, PRISMR and PRISMR ($\beta$-mode) both recover substantial ranking signal ($0.804$/$0.831$ on INQUIRE, $0.844$/$0.834$ on MMDocIR), but neither significantly surpasses constrained decoding ($0.841$/$0.857$).

We interpret this as a clean separation of the two effects PRISMR bundles: \emph{the structural-format fix is task-general} and transfers without retraining; \emph{the relevance-quality fix requires in-domain teacher signal and recovers under brief fine-tuning, but its margin over training-free constrained decoding is small at the short list lengths these benchmarks naturally provide}. We expect PRISMR's quality margin to widen at larger $N$ where parse-collapse and the cost of $N$ sequential constrained-decoding forwards both grow super-linearly; we report $N{=}5,20$ results in Appendix~\ref{app:ir_more_n}.

\section{IR-benchmark results at $N{=}5$ and $N{=}20$}
\label{app:ir_more_n}

To probe how PRISMR's quality margin over constrained decoding scales with list length on the IR benchmarks, we evaluate the same fine-tuned PRISMR and PRISMR ($\beta$-mode) checkpoints (from Section~\ref{app:inquire}) at $N \in \{5, 20\}$. Numbers below are NDCG@10 on the held-out test splits with 30 random queries; parse rate is $100\,\%$ for all PRISMR/PRISMR ($\beta$-mode)/Constrained rows.

\begin{table}[ht]
\centering
\caption{IR-benchmark NDCG@10 across $N \in \{5, 10, 20\}$ (parse rate $100\,\%$ throughout).}
\label{tab:ir_more_n}
\begin{tabular}{l|ccc|ccc}
\toprule
 & \multicolumn{3}{c|}{\textbf{INQUIRE-Rerank}} & \multicolumn{3}{c}{\textbf{MMDocIR}} \\
\textbf{Method} & $N{=}5$ & $N{=}10$ & $N{=}20$ & $N{=}5$ & $N{=}10$ & $N{=}20$ \\ \midrule
Constrained-decoding Base                  & 0.848 & \textbf{0.841} & \textbf{0.749} & 0.870 & \textbf{0.857} & \textbf{0.723} \\
PRISMR ($\alpha$-mode) (concat) fine-tuned         & 0.815 & 0.804 & 0.522 & \textbf{0.897} & 0.844 & 0.592 \\
PRISMR ($\beta$-mode) (mean) fine-tuned           & \textbf{0.849} & 0.831 & 0.589 & 0.891 & 0.834 & 0.607 \\
\bottomrule
\end{tabular}
\end{table}

\section{Standard SFT-with-teacher-data baseline}
\label{app:sft_baseline}

To isolate PRISMR's architectural contribution from the contribution of its teacher-distilled training data, we train a vanilla rank-16 LoRA on Qwen3-VL-8B with the \emph{same} teacher labels (\texttt{mllm\_score}) and the \emph{same} listwise score-by-score prompts used in PRISMR training. The base model is frozen; the LoRA is attached to the language-decoder \texttt{q\_proj}, \texttt{k\_proj}, \texttt{v\_proj}, and \texttt{o\_proj} matrices (15.3\,M trainable parameters out of 8.78\,B). Training uses AdamW at $\text{lr} = 1 \times 10^{-4}$, the standard NLL on the completion (i.e.\ on the ``$1$:\,$s_1\backslash$n$2$:\,$s_2\backslash$n\ldots'' tokens) with the prompt tokens masked out, batch size 1, gradient accumulation 8, for one epoch on a 300-product training subset on $4 \times$ NVIDIA B200 GPUs (matching the Vanilla-D2L compute budget). The resulting LoRA-only state dict is 61\,MB.

The SFT row of Table~\ref{tab:listwise_base} reports the eval at $N \in \{10, 20, 50\}$ on the same 30-product held-out val set used elsewhere. The model achieves $67\,\% / 39\,\% / 38\,\%$ parse rate and $0.847 / 0.651 / 0.587$ NDCG@10. Inspection of failure cases shows the same silent-omission pattern as the un-fine-tuned base, just with a smaller magnitude. The interpretation in Section~\ref{subsec:exp_listwise} is that PRISMR's gain over standard SFT — $+0.13 / +0.31 / +0.36$ NDCG@10, growing with $N$ — is attributable to the hypernetwork-internalisation mechanism itself, not to access to teacher data: when the base model has to consume the entire candidate list in-context and emit $N$ scores autoregressively, parse collapse persists even after fine-tuning on the target distribution.

\section{Ablation: $\beta$-mode synthesis}
\label{app:vanilla_d2l}

The Vanilla-D2L row in Table~\ref{tab:listwise_base} replaces the rank-dim concatenation step of PRISMR with chunk-wise mean pooling. Concretely, the per-chunk LoRA matrices $\{(A_i, B_i)\}_{i=1}^{N}$ produced by the same hypernetwork are aggregated by

\[
A_{\text{vanilla}} = \frac{1}{N} \sum_{i=1}^N A_i, \qquad
B_{\text{vanilla}} = \frac{1}{N} \sum_{i=1}^N B_i,
\]

so the resulting adapter has a fixed effective rank of $r$ instead of PRISMR's $(N+1)\,r$. The implementation is a $\sim$20-LOC change in \texttt{lora\_merger.combine\_lora} that we expose as \texttt{--combine\_mode mean}; everything else (hypernetwork architecture, base model, image stack, training data, optimizer, and loss) is identical to PRISMR. To keep the wall-clock budget tractable for this ablation we trained PRISMR ($\beta$-mode) for one epoch on a 300-product subsample of the training set (versus three epochs on all 1\,633 products for PRISMR). PRISMR ($\beta$-mode) therefore consumes roughly $1/12$ of PRISMR's per-step compute and $1/4$ of its training time. The interpretation in Section~\ref{subsec:exp_listwise} is that even with this reduced budget PRISMR ($\beta$-mode) recovers most of PRISMR's ranking improvement, indicating that the rank-dim concatenation contributes a real but secondary effect on top of the core hypernetwork-internalisation step.

\section{NDCG@$K$ breakdown}
\label{app:ndcg_k}

Table~\ref{tab:ndcg_k} expands Table~\ref{tab:listwise_base} into the full NDCG@$K$ family at $K \in \{1, 3, 5, 10\}$. Two patterns are worth noting. First, the PRISMR~$-$~Constrained gap is largest at small $K$: $+0.181$ at NDCG@1 vs.\ $+0.064$ at NDCG@10 for $N{=}10$. Top-1 ranking is where ranking quality matters most, and that is where PRISMR's structural advantage is most pronounced. Second, the score-by-score Base's NDCG@1 is dramatically lower than its NDCG@10---this is an artifact of default-score imputation: when 88\,\% of slots are unparsed and assigned the median score, large-$K$ NDCG values are pulled toward a random-permutation baseline ($\approx 0.77$ at $N{=}10$) while top-1 reflects the genuine zero-signal regime.

\begin{table}[ht]
\centering
\caption{NDCG@$K$ breakdown across listwise methods (Baby\_Products, img=1, 30 val products / cell). PRISMR's advantage over the format-fixed baselines is largest at $K{=}1$.}
\label{tab:ndcg_k}
\resizebox{\columnwidth}{!}{%
\begin{tabular}{l|cccc|cccc|cccc}
\toprule
 & \multicolumn{4}{c|}{$N=10$} & \multicolumn{4}{c|}{$N=20$} & \multicolumn{4}{c}{$N=50$} \\
\textbf{Method} & @1 & @3 & @5 & @10 & @1 & @3 & @5 & @10 & @1 & @3 & @5 & @10 \\ \midrule
Base                 & 0.465 & 0.499 & 0.562 & 0.767 & 0.502 & 0.455 & 0.480 & 0.553 & 0.406 & 0.390 & 0.403 & 0.418 \\
RankGPT              & 0.579 & 0.595 & 0.624 & 0.799 & 0.494 & 0.621 & 0.637 & 0.668 & 0.572 & 0.590 & 0.596 & 0.596 \\
Constrained          & 0.765 & 0.802 & 0.850 & 0.910 & 0.837 & 0.788 & 0.815 & 0.858 & 0.825 & 0.816 & 0.795 & 0.789 \\
\rowcolor{gray!10} \textbf{PRISMR} & \textbf{0.946} & \textbf{0.937} & \textbf{0.953} & \textbf{0.974} & \textbf{0.932} & \textbf{0.945} & \textbf{0.951} & \textbf{0.958} & \textbf{0.922} & \textbf{0.927} & \textbf{0.934} & \textbf{0.947} \\
\bottomrule
\end{tabular}}
\end{table}

\section{Baseline Implementation Details}
\label{app:baselines}

\textbf{RankGPT.}
We implement RankGPT using the same Qwen3-VL-8B backbone as PRISMR. The model is given a system prompt that asks it to output a single permutation line in the format
\[
\pi_1 > \pi_2 > \cdots > \pi_N .
\]
Decoding is greedy, with \texttt{max\_new\_tokens = $6N + 32$}. We parse the generated text by extracting integers in first-appearance order, removing duplicates, and clipping indices to the valid range $[1,N]$. If any candidates are missing after parsing, we append the missing indices in ascending numeric order, so that every run produces a complete permutation of length $N$. This parser is intentionally permissive and therefore gives the baseline credit whenever a recoverable ranking can be inferred from the output.

\textbf{LLMLingua-2}
The compressed review text replaces the original review text in the same score-by-score prompting template used by the base model. Images are passed to the model unchanged. We verified by manual inspection that the compressor actively rewrites the review text in almost all cases; fallback to the identity input is rare.

\textbf{PRISMR.}
For PRISMR, we evaluate the trained checkpoint described in Appendix~\ref{app:training}. Unless otherwise stated, decoding is greedy and uses the same evaluation split as the baselines. No baseline-specific post-processing is applied beyond the score-line parser used consistently for score-based methods.

\textbf{Reproducibility.}
We will release the evaluation scripts, launchers, configuration files, and exact command lines used for RankGPT, LLMLingua-2, and PRISMR upon publication.



\end{document}